\documentclass[11pt,letterpaper]{article}
\usepackage{coling2016}
\usepackage{times}
\usepackage{latexsym}

\usepackage{amsmath}
\usepackage{amssymb,amsmath,epsfig}
\usepackage{bbm}
\usepackage{cprotect}
\usepackage{tikz}
\usepackage{tikz-qtree}
\usepackage{graphics}
\usepackage{graphicx}
\usepackage{grffile}
\usepackage[position=b]{subcaption}

\usepackage{color}
\usepackage{tablefootnote}
\usepackage{calc}
\usepackage{array}
\usepackage[export]{adjustbox}
\usepackage{url}
\usepackage{xspace}
\makeatletter
\newcommand{\pushright}[1]{\ifmeasuring@#1\else\omit\hfill$\displaystyle#1$\fi\ignorespaces}
\newcommand{\pushleft}[1]{\ifmeasuring@#1\else\omit$\displaystyle#1$\hfill\fi\ignorespaces}
\makeatother

\providecommand{\card}[1]{\lvert#1\rvert}  % cardinality |x|

\newcommand{\Prob}{\ensuremath{\mathrm{Pr}}\xspace}
\newcommand{\expect}{\ensuremath{\mathbb{E}}}

\newcommand{\mtxa}{\ensuremath{\mathbf{A}}\xspace}
\newcommand{\veca}{\ensuremath{\mathbf{A}}\xspace}
\newcommand{\vece}{\ensuremath{\mathbf{e}}\xspace}
\newcommand{\veci}{\ensuremath{\mathbf{i}}\xspace}
\newcommand{\vecc}{\ensuremath{\mathbf{c}}\xspace}
\newcommand{\vech}{\ensuremath{\mathbf{h}}\xspace}
\newcommand{\veco}{\ensuremath{\mathbf{o}}\xspace}
\newcommand{\vecu}{\ensuremath{\mathbf{u}}\xspace}
\newcommand{\vecf}{\ensuremath{\mathbf{f}}\xspace}
\newcommand{\vecx}{\ensuremath{\mathbf{x}}\xspace}

\newcommand{\defeq}{\ensuremath{\stackrel{\Delta}{=}}\xspace}

\title{Textual Entailment with Structured Attentions and Composition}

\author{Kai Zhao \and Liang Huang \and Mingbo Ma \\
School of Electrical Engineering and Computer Science \\ 
Oregon State University \\ Corvallis, Oregon, USA \\
{\tt \{kzhao.hf, lianghuang.sh, cosmmb\}@gmail.com}}

\date{}

\begin{document}

\maketitle

\vspace{-0.35in}
\begin{abstract}
Deep learning techniques are increasingly popular 
in the textual entailment task, % is witnessing a burgeoning
%interest in leveraging the generalization power of 
%deep learning techniques
overcoming the fragility of traditional discrete models with hard alignments and logics.
In particular, the recently proposed attention models 
\cite{rocktaschel2015reasoning,wang2015learning} achieves state-of-the-art accuracy by
computing soft word alignments between %words
the premise and hypothesis sentences.
However, there remains a major limitation:
this line of work completely ignores syntax and recursion,
%outperforming traditional models with sparse features. % on a large dataset.
%However, syntactic trees, 
which is helpful in many traditional efforts.
%is completely ignored in this line of work.
We show that it is beneficial to extend the attention model 
to tree nodes between premise and hypothesis.
More importantly, this subtree-level
attention reveals information about entailment relation.
We study the recursive composition of this subtree-level entailment relation,
%and propose to combine the attention calculation and the entailment
%composition,
which can be viewed as a soft version of 
the Natural Logic framework
\cite{maccartney2009extended}.
Experiments show that our structured attention and
entailment composition model can correctly identify and infer
entailment relations from the bottom up,
and bring significant improvements in accuracy.
\end{abstract}

\section{Introduction}
\label{sec:intro}
%!TEX root = main.tex

Automatically recognizing sentence entailment relations between 
a pair of sentences has long been believed to be an ideal testbed
for discrete approaches using alignments and 
rigid logic inferences
\cite{zanzotto2009machine,maccartney2009extended,wang2010probabilistic,watanabe2012latent,tian2014logical,filice2015structural}.
All of these methods are based on sparse features, 
making them brittle for unseen phrases and sentences.

Recent advances in deep learning reveal 
another promising direction to solve this problem. 
Instead of discrete features and logics, 
continuous representation of the sentence is more 
robust to unseen features  
without sacrificing performance \cite{bowman2015large}. 
In particular, the attention model based on 
LSTM can successfully
identify the word-by-word correspondences between the two sentences 
that lead to entailment or contradiction, 
which makes the entailment relation inference more focused on
local information and less vulnerable to misleading information
from other parts of the sentence \cite{rocktaschel2015reasoning,wang2015learning}. 

However, conventional neural attention models for 
entailment recognition problem treat sentences as sequences, 
ignoring the fact that sentences are formed 
from the bottom up with syntactic tree structures, 
which inherently associate with the semantic meanings. 
Thus, using the tree structure of the sentences will be beneficial 
in inducing the entailment relations between parts of the two sentences, 
and then further improving the sentence-level 
entailment relation classification \cite{watanabe2012latent}.

Furthermore, as \newcite{maccartney2009extended} point
out, the entailment relation between sentences is modular,
and can be modeled as the composition of
subtree-level entailment relations.
These subtree-level entailment relations are induced
by comparing subtrees between the two sentences,
which are by nature a perfect match to be modeled
by the attention model over trees.

In this paper we propose a recursive neural network model 
that calculates the attentions following the tree structures, 
which helps determine entailment relations between parts of the sentences. 
We model the entailment relation 
with a continuous representation.%, and then
%use this entailment information on subtrees 
%to infer the entailment relations 
%on the parent node as a composition operation. 
The relation representations of non-leaf nodes
are recursively computed by composing their children's 
relations.
This approach can be viewed as a {\it soft} version of 
Natural Logic \cite{maccartney2009extended} 
for neural models, 
and can make the recognized entailment relation easier to interpret.

We make the following contributions:
\begin{enumerate}
\item We adapt the sequence attention model to the tree structure.
This attention model directly works on meaning representations
of nodes in the syntactic trees,
and provides a more precise guidance for 
subtree-level entailment relation inference. (Section~\ref{sec:att})
\item We propose a continuous representation for entailment
relation that is specially designed for entailment composition
over trees. This entailment relation representation
is recursively composed to induce the overall
entailment relation, and is easy to interpreted.
(Section~\ref{sec:ent})
\item Inspired by the forward and reverse
alignment technique in machine translation,
we propose dual-attention that 
considers both the premise-to-hypothesis
and hypothesis-to-premise directions, which makes the
attention more robust to confusing alignments. (Section~\ref{sec:dual})
\item Experiments show that our model brings significant 
performance boost based on a Tree-LSTM model.
Our dual-attention can provide superior guidance for the entailment relation inference
(Figure~\ref{fig:align-example}).
The entailment composition follows the intuition of Nature Logic
and can provide a vivid illustration of how the final entailment
 conclusion
is formed from bottom up (Figure~\ref{fig:ent-example}).
(Section~\ref{sec:exp})  
\end{enumerate}

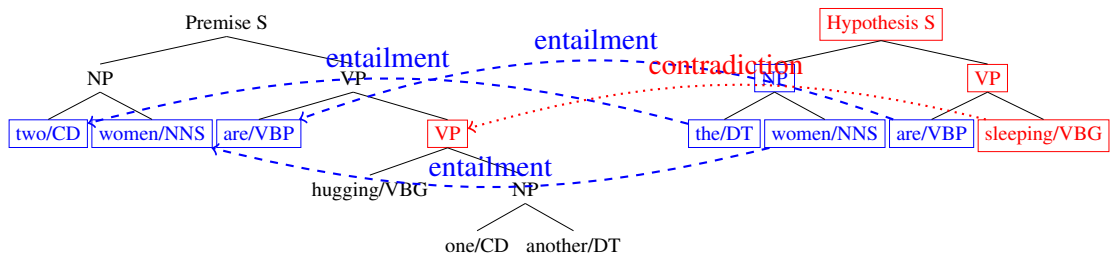
\begin{figure*}
\begin{center}
\begin{tikzpicture}[scale=0.7]
\begin{scope}[frontier/.style={distance from root=120pt}]
\Tree[.{Premise S} [.NP \node[draw,color=blue](p1){two/CD}; \node[draw,color=blue](p2){women/NNS}; ] [.VP \node[draw,color=blue](p3){are/VBP}; [.\node[draw,color=red](p4){VP}; hugging/VBG [.NP one/CD another/DT ] ] ] ]
\end{scope}
\begin{scope}[xshift=350pt,frontier/.style={distance from root=120pt}]
\Tree[.\node[draw,color=red]{Hypothesis S}; [.\node[draw,color=blue]{NP}; \node[draw,color=blue](h1){the/DT}; \node[draw,color=blue](h2){women/NNS}; ] [.\node[draw,color=red]{VP}; \node[draw,color=blue](h3){are/VBP}; \node[draw,color=red](h4){sleeping/VBG}; ] ]
\end{scope}

\begin{scope}[dashed]
\draw[thick,->,color=blue] (h1) to [bend right=15] node[above] {entailment}  (p1);
\draw[thick,->,color=blue] (h2) to [bend left=14] node[above] {entailment}  (p2);
\draw[thick,->,color=blue] (h3) to [bend right=20] node[above] {entailment}  (p3);
\draw[thick,dotted,->,color=red] (h4) to [bend right=15] node[above] {contradiction}  (p4);
\end{scope}
\end{tikzpicture}
\end{center}
\caption{Exemplary trees for the premise sentence 
``two women are hugging one another'' and 
the hypothesis sentence ``the women are sleeping''.
The syntactic labels (NP, VP, CD, etc.) are not used in the model.
The dashed and dotted lines show the lowest level of 
alignments from the hypothesis tree nodes
to the premise tree nodes.
The blue dashed lines mark the entailment relations,
and the red dotted line marks the contradiction relation.
In the hypothesis tree, tree nodes in blue squares
are identified to be entailment from the premise,
and nodes in red squared are identified to contradicts
the premise.
By composing these relations from the bottom up,
we reach a conclusion that the sentence-level entailment
relation is contradiction.
Please also refer to Figure~\ref{fig:ent-example}
for real examples taken from our experiments.
\label{fig:egtree}}
\end{figure*}

\section{Structured Attentions \& Entailment Composition}
\label{sec:model}
%!TEX root = main.tex

Here we first give an overview and formalization of our model, 
and then describe its components.

\subsection{Formalization}

We assume both the premise tree and the hypothesis tree are 
binarized.

We use the premise tree and hypothesis tree 
in Figure~\ref{fig:egtree} to demonstrate the process 
of our approach.
The premise sentence is ``two women are hugging one another'',
and the hypothesis sentence is ``the women are sleeping''.

Following the traditional approaches \cite{maccartney2009extended,watanabe2012latent}, 
we first find the alignments from hypothesis
tree nodes to premise tree nodes
(i.e., the dashed or dotted curves in Figure~\ref{fig:egtree}).
Then we explore inducing the sentence-level 
entailment relations by 
1) first computing the entailment relation
at each node of the hypothesis tree based on the alignments, 
and then 2) composing the entailment relations
at the internal hypothesis nodes from bottom up
to the root in a recursive way.
Our model resembles the work of Natural Logic
\cite{maccartney2009extended} in the spirit 
that the entailment relation is inferred {\it modularly},
and composed {\it recursively}.

We formalize this entailment task as a structured prediction problem
similar to \newcite{mnih2014recurrent}, \newcite{ba2015learning}, and \newcite{xu2015show}.
The inputs are two trees: premise tree $P$,
and hypothesis tree $Q$. The goal is to predict a label
$y \in \{\text{contradiction}, \text{neutral}, \text{entailment}\}$.
Note that although the output label $y$ is not structured,
we can still consider the problem as a structured prediction problem,
because: 1) the input is a pair of trees; and 2)
the internal alignments are structured.

More formally, we aim to minimize the negative 
log likelihood of the 
gold label given the two trees. The objective
can be written in the online fashion as:
\vspace{-0.2cm}
\begin{align*}
\ell = & -\log\Prob(y|P, Q) = -\log \sum_{\mtxa} \Prob(y, \mtxa|P, Q)\\
= & -\log \sum_{\mtxa} \Prob(\mtxa|P, Q)\cdot \Prob(y|\mtxa, P, Q)
=  -\log \expect_{\Prob(\mtxa|P, Q)} [\Prob(y|\mtxa, P, Q)],
\end{align*}\\[-0.5cm]
where the structured latent variable
$\mtxa\in \{0, 1\}^{|Q|\times |P|}$ represents an alignment.
$\card{\cdot}$ is the number of nodes in the tree.
$\mtxa_{ij}=1$ if and only if node $i$ in $Q$ is aligned to node $j$
in $P$, otherwise $\mtxa_{ij}=0$.

%Our structured entailment model accomplish these two objectives
%in one step of calculation 
%and will be discussed in Section~\ref{sec:entailment}.

However, enumerating over all possible alignments $\mtxa$ takes exponential
time, we need to efficiently approximate the above log expectation.

\begin{figure*}
\raisebox{1.02in}{
\begin{subfigure}[t]{0.48\textwidth}
%\begin{center}
$\quad\ \ \ \frac{1}{3}\cdot$\raisebox{-.5\height}{\includegraphics[width=0.2\textwidth]{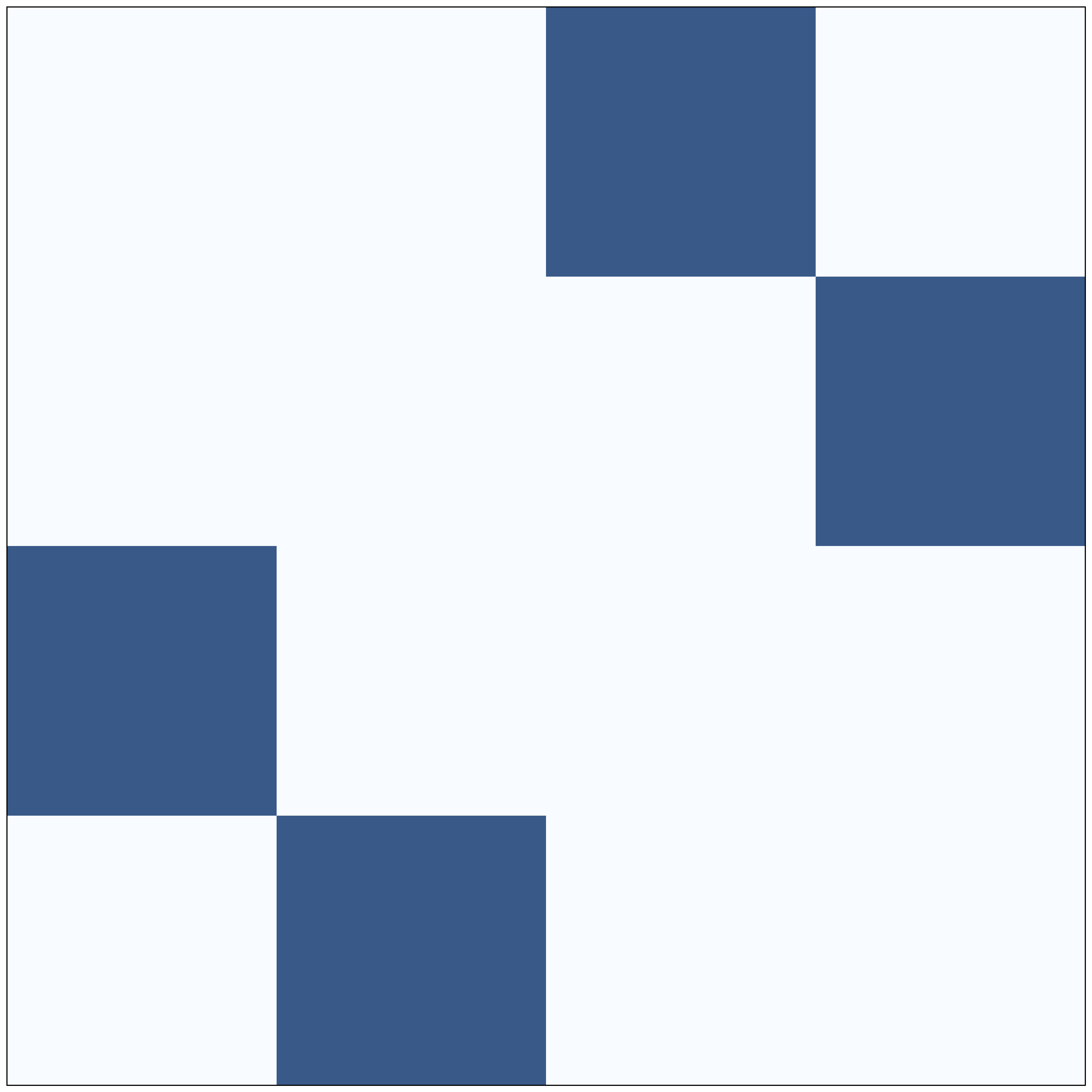}}
$\quad+\quad \frac{1}{2}\cdot$\raisebox{-.5\height}{\includegraphics[width=0.2\textwidth]{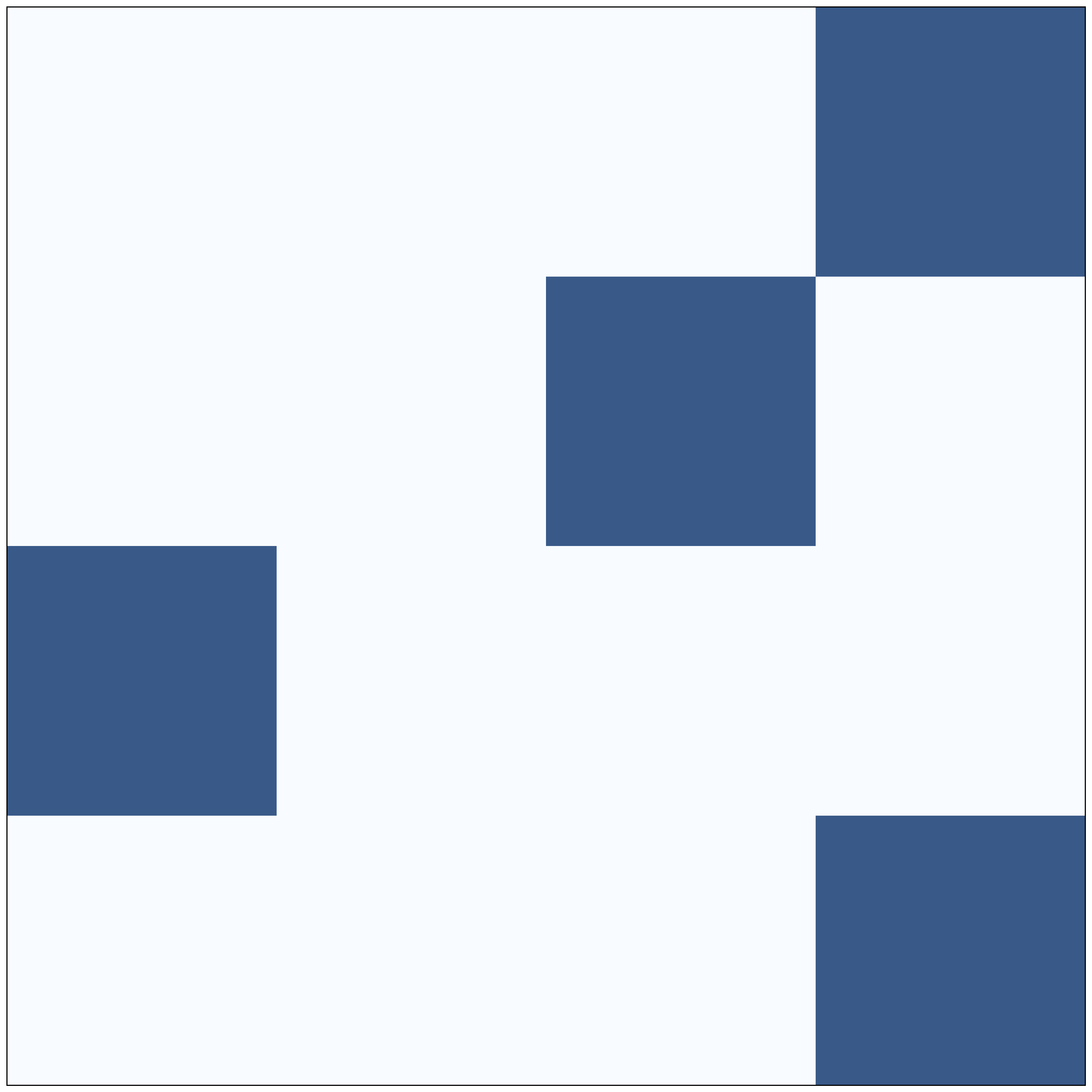}}\\[0.1in]
$\quad+\quad \frac{1}{6}\cdot$\raisebox{-.5\height}{\includegraphics[width=0.2\textwidth]{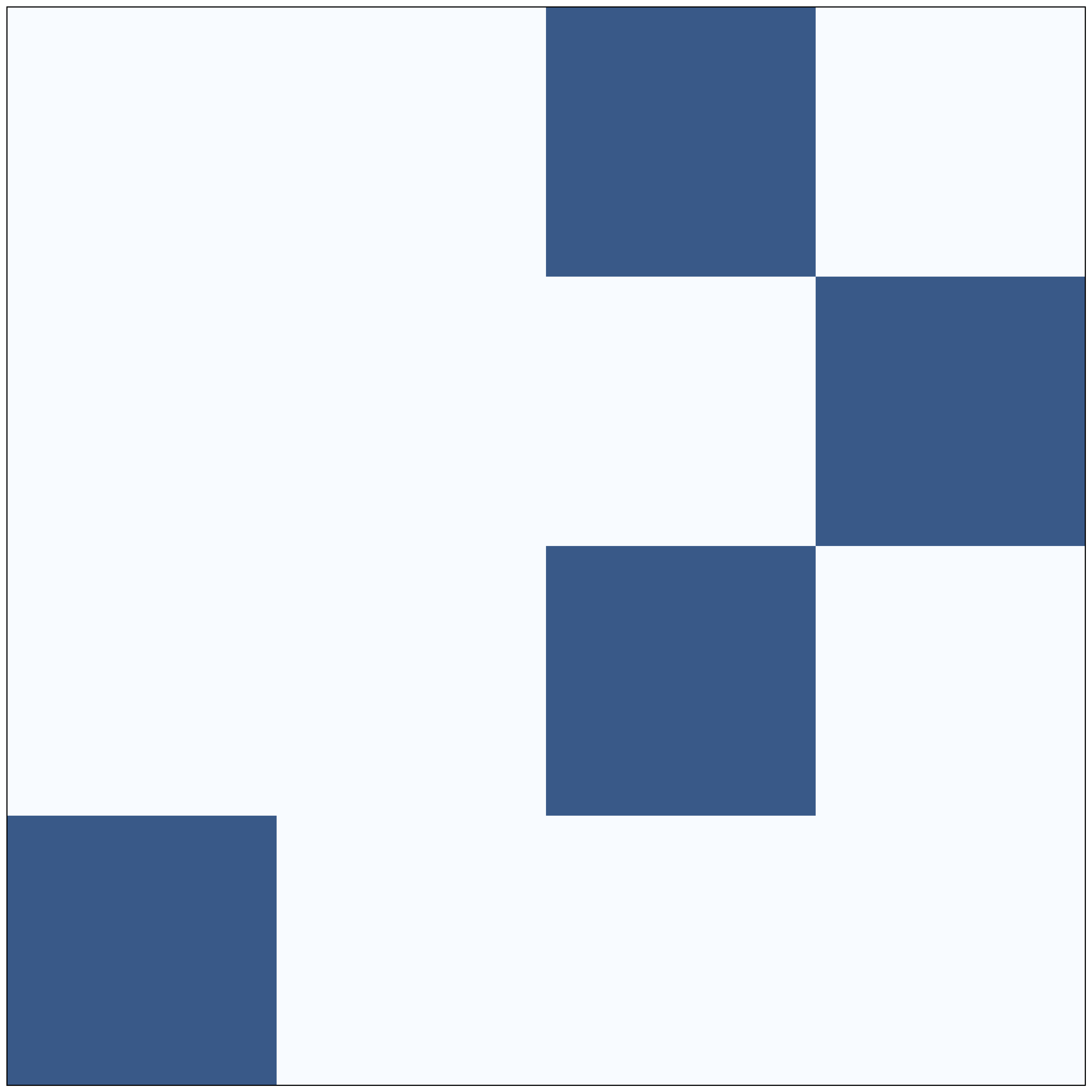}}
$\quad=\quad\ \ \  $\raisebox{-.5\height}{\includegraphics[width=0.2\textwidth]{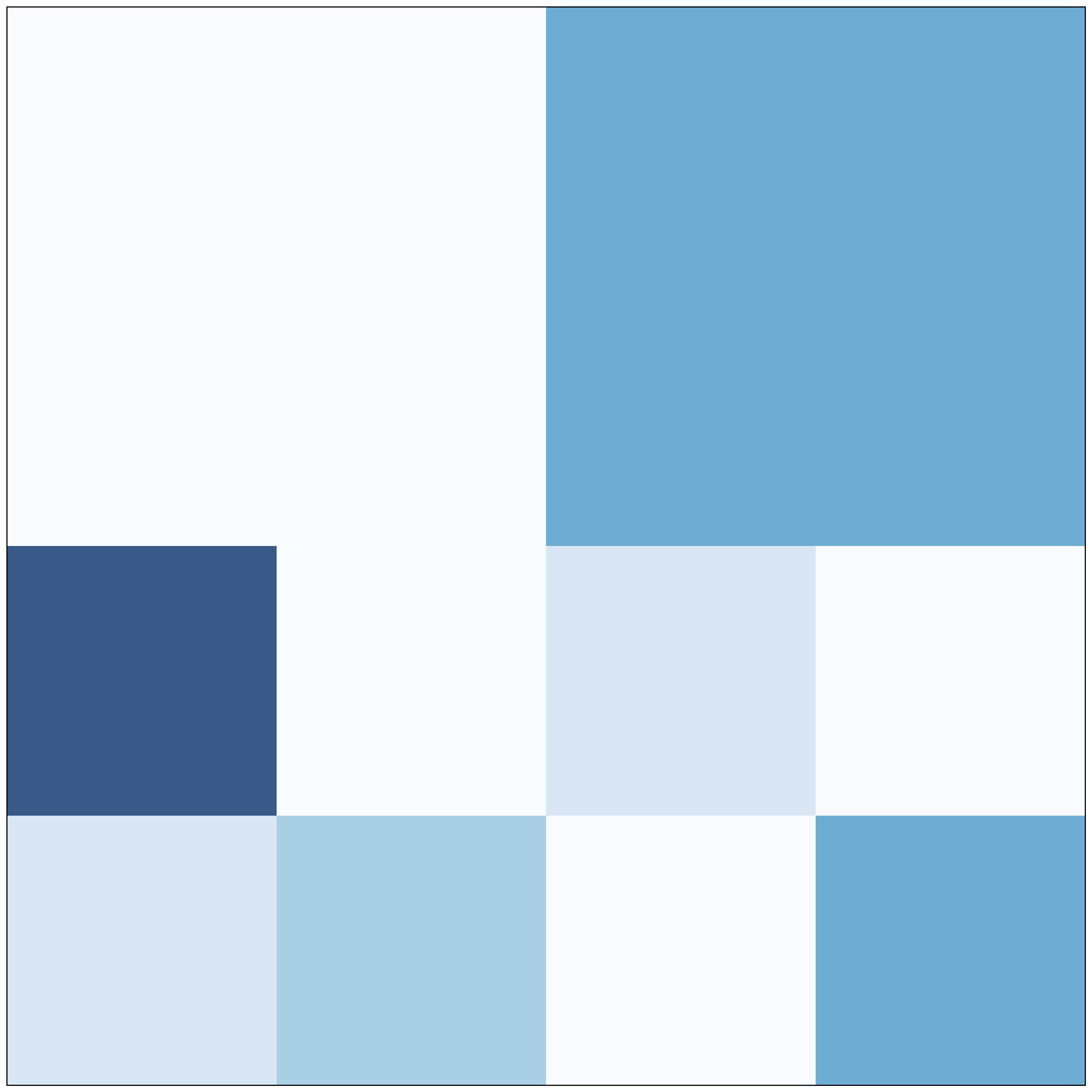}}
%\end{center}
\caption{Expected alignment over 3 alignments with probability distribution of $(\frac{1}{3}, \frac{1}{2}, \frac{1}{6})$.
Each alignment is a matrix of $\{0, 1\}^{4\times 4}$,
and the expected alignment is a matrix of $\mathbb{R}^{4\times 4}$.
\label{fig:exptalign}}
\end{subfigure}
}
$\ $
\begin{subfigure}[t]{0.48\textwidth}
%\begin{center}
\includegraphics[width=0.9\textwidth]{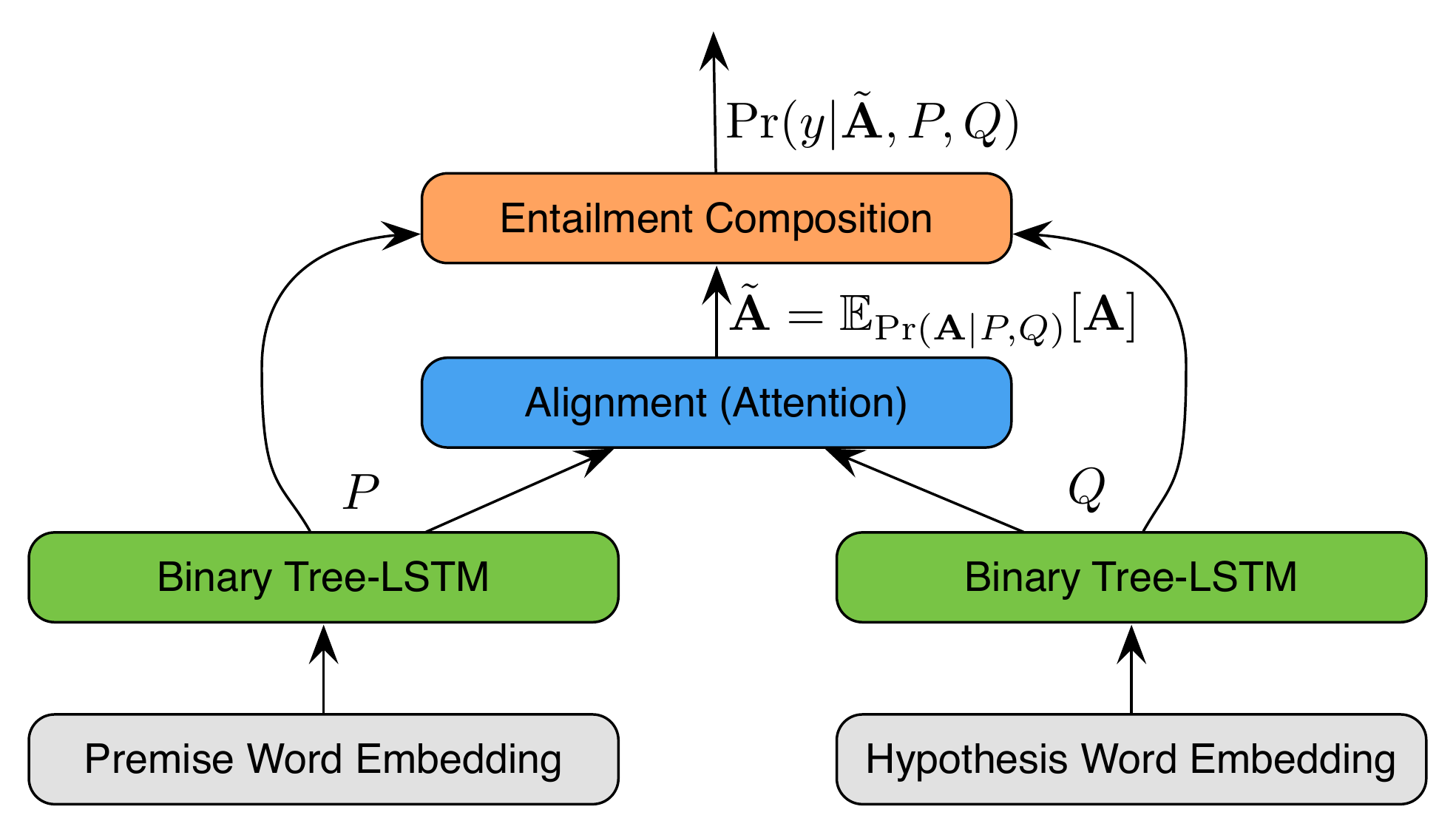}
%\end{center}
\caption{Architecture of our neural network.
Layers with the same color share the same parameters.
%In this network, first the Binary-Tree LSTM
%computes the meaning representations for each
%tree node. Then the Alignment module
The alignment module
calculates the expected alignment $\tilde{\mtxa}$.
%based on the node meaning representation.
The Entailment Composition module
infers the final entailment $y$.%relation along the 
%hypothesis tree from bottom up
%using the node meaning representation
%and the expected alignment.
\label{fig:arch}}
\end{subfigure}
\vspace{-0.2cm}
\caption{Expected alignments calculation (a) and overview of the network architecture (b).}
\vspace{-0.5cm}
\end{figure*}

Fortunately, as \newcite{xu2015show} point out, as long as the calculation
$\Prob(y|\mtxa, P, Q)$ only consists of linear calculation,
simple nonlinearities like $\tanh$, and softmax, we can 
have following simplification via first-order Taylor
approximation:
%\begin{align*}
\[
\ell =  -\log \expect_{\Prob(\mtxa|P, Q)} [\Prob(y|\mtxa, P, Q)]
\approx  -\log \Prob(y|\expect_{\Prob(\mtxa|P, Q)}[\mtxa], P, Q)],
%\end{align*}
\]\\[-0.7cm]
which means instead of enumerating over all alignments 
and calculating the label probability for each alignment,
we can use the label probability for the expected alignment 
as an approximation:\footnote{
We use bold letter, $\mtxa$,
for binary alignments, and tilde version, $\tilde{\mtxa}$, for the 
expected alignments in the real number space.
}
\begin{equation}
\tilde{\mtxa} \defeq \expect_{\Prob(\mtxa|P, Q)}[\mtxa] \quad \in \mathbb{R}^{\card{Q}\times \card{P}}
\label{eq:expatt}
\end{equation}\\[-0.7cm]
Figure~\ref{fig:exptalign} shows an example of expected
alignment calculation.
The objective is simplified to
\begin{equation}
\ell \approx -\log \Prob(y|\tilde{\mtxa}, P, Q).
\label{eq:newobj}
\end{equation}\\[-1.1cm]

\captionsetup[subfigure]{position=b}

With this observation, we split our calculation into two steps
as the top two modules in Figure~\ref{fig:arch}.
First in the Alignment module,
we calculate the expected alignments 
$\tilde{\mtxa}$ using Equation~\ref{eq:expatt} 
(Section~\ref{sec:att}).
Then we calculate the node-wise
entailment relation, propagate and compose the relation from
bottom up to find out the final entailment relation 
(Equation~\ref{eq:newobj})
in the Entailment Composition module
(Section~\ref{sec:ent}).
Both of these two modules rely on the composition of tree node meaning
representations (Section~\ref{sec:treelstm}).

\iffalse
\begin{figure}
\vspace{-0.4cm}
\begin{center}
\includegraphics[width=0.48\textwidth]{figures/arch}
\end{center}
\caption{Architecture of our neural network.
Layers with the same color share the same parameters.
In this network, first the Binary-Tree LSTM
computes the meaning representations for each
tree node. Then the Alignment module
calculates the expected alignment
based on the node meaning representation.
Finally the Entailment Composition module
infers the entailment relation along the 
hypothesis tree from bottom up
using the node meaning representation
and the expected alignment.
\label{fig:arch}}
%\vspace{-0.4cm}
\end{figure}
\fi

%%%%%%%%%%%%%%%%%%%%%%%%%%%%%%%%%%%%%%%%%%%%%%%%%%%%%%
\subsection{Attention over Tree Nodes}
\label{sec:att}
First we calculate the expected alignments
 $\tilde{\mtxa}$ between the hypothesis $Q$ and the premise $P$ (Equation~\ref{eq:expatt}):
\[\tilde{\mtxa} = \expect_{\Prob(\mtxa|P, Q)}[\mtxa].\]

To simplify the calculation, we further
approximate the global (binary) alignment $\mtxa$
to be consisted of the alignment 
$\mtxa_i \in \{0, 1\}^{1\times |P|}$ 
of each tree node $i\in Q$
independently. $\mtxa_i$ is the $i$th row of $\mtxa$:
\[\mtxa = [\mtxa_1^T;\mtxa_2^T;\dots;\mtxa_{\card{Q}}^T]^T,\]\vspace{-0.2in}
\[\Prob(\mtxa|P, Q) = \prod_i^{\card{Q}} \Prob(\mtxa_i|P, Q).\]
$\Prob(\mtxa_{i,j}=1|P, Q)$ is the probability 
of the node $i\in Q$ being aligned to node $j\in P$,
which is defined as:
\begin{equation}
\Prob(\mtxa_{i,j}=1|P, Q) \defeq 
\frac{\exp(T_{2k, 1}([\vech_i;\vech_j]))}{\sum_k\exp(T_{2k, 1}([\vech_i;\vech_k]))}.\label{eq:att}
\end{equation}
$\vech_i, \vech_j \in \mathbb{R}^k$ are vectors representing the semantic meanings of node $i$, $j$, respectively,
whose calculation will be described in Section~\ref{sec:treelstm}.
$T_{2k,1}$ is an affine transformation 
from $\mathbb{R}^{2k}$ to $\mathbb{R}$.
This formulation essentially is equivalent to 
the widely used attention
calculation in neural networks \cite{bahdanau2014neural}, i.e.,
for each node $i\in Q$, we find the relevant 
nodes $j\in P$ and use the softmax of the relevances 
as a probability distribution. In the rest of the paper,
we use ``expected alignment'' and ``attention'' interchangeably.

The expected alignment of node $i$ being aligned to node $j$, 
by definition, is:
\[
\tilde{\mtxa}_{i,j} = \Prob(\mtxa_{i,j}=1|P,Q) \cdot 1 = \Prob(\mtxa_{i,j}=1|P,Q).
\]
%Similarly we use $\tilde{\mtxa}_{i,j}$ to denote the probability 
%that node $i$ being aligned to node $j$:
%\[
%\tilde{\mtxa}_{i,j} = \sum_{\mtxa_i} \Prob(\mtxa_i|P, Q) \mtxa_i\quad \in \mathbb{R}^{1\times\card{P}}.
%\]

%%%%%%%%%%%%%%%%%%%%%%%%%%%%%%%%%%%%%%%%%%%%%%%%%%%%%%
\subsection{Entailment Composition}
\label{sec:ent}

Now we can calculate the entailment relation at each tree node
and propagate the entailment relation following
the hypothesis tree from bottom up, assuming the expected alignment
is given (Equation~\ref{eq:newobj}):
\[\ell \approx -\log \Prob(y|\tilde{\mtxa}, P, Q).
\]

Let vector $\vece_i \in \mathbb{R}^r$ denote the entailment relation in a 
latent relation space
at hypothesis tree node $i\in Q$. 
At the root of the hypothesis tree.
We can induce the final entailment relation from entailment relation vector
 $\vece_{\text{root}}$. We use a simple $\tanh$ layer
to project the entailment relation to the 3 relations
defined in the task, and use a softmax layer to 
calculate the probability for each relation:
\[
\Prob(y|\tilde{\mtxa}, P, Q) = \text{softmax}(\tanh (T_{r, 3}( \vece_{\text{root}}))).
\]

At each hypothesis node $i$, $\vece_i$ is calculated recursively given
the meaning representation at this tree node $\vech_i$,
the meaning representation of every node in the premise tree $\vech_j, j\in P$,
and the entailment from $i$'s children, $\vece_{i, 1}, \vece_{i, 2}$:
\begin{equation}
\vece_i 
= f_{\text{rel}}([\vech_i; \sum_{j\in P}\tilde{\mtxa}_{i,j} \vech_j], \vece_{i, 1}, \vece_{i, 2})
\label{eq:relcomp}
\end{equation}

\begin{figure}
%\begin{center}
\centering
\begin{subfigure}[b]{0.49\textwidth}
\centering
\includegraphics[width=0.65\textwidth]{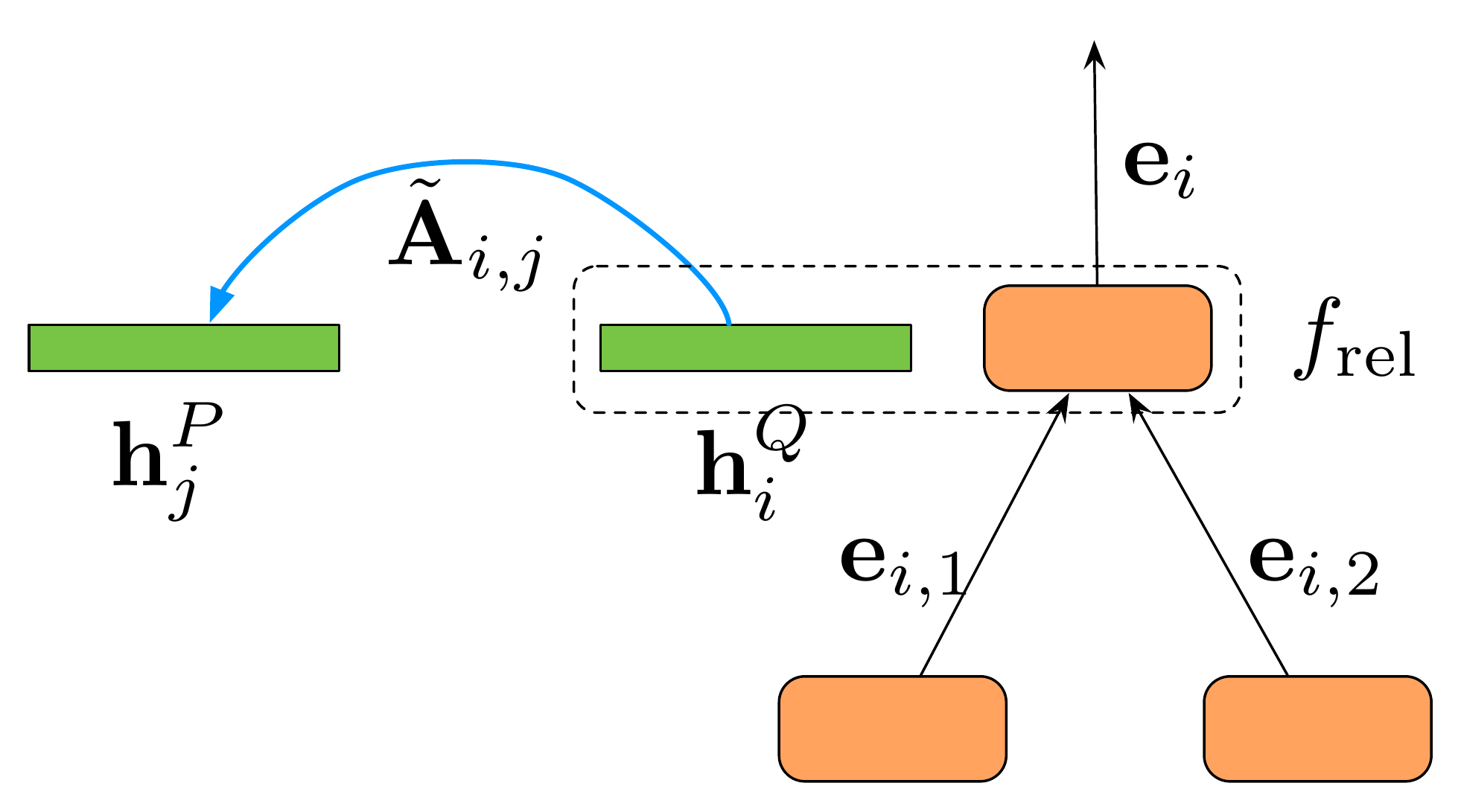}
%\end{center}
\caption{Entailment composition at hypothesis tree node $i$.
The composition is based on the meaning representation
of current node $\vech_i^Q$, 
the expected alignment for node $i$, $\tilde{\mtxa}_i$,
expected meaning representation of aligned premise
tree node $\sum_{j\in P}\tilde{\mtxa}_{i,j} h_j^P$, 
and known entailment relations from children nodes 
$\vece_{i, 1}, \vece_{i,2}$.
\label{fig:entnode}
}
\end{subfigure} 
$\ $
\begin{subfigure}[b]{0.49\textwidth}
\begin{center}
$\Bigg($\raisebox{-.4\height}{\includegraphics[width=0.16\textwidth]{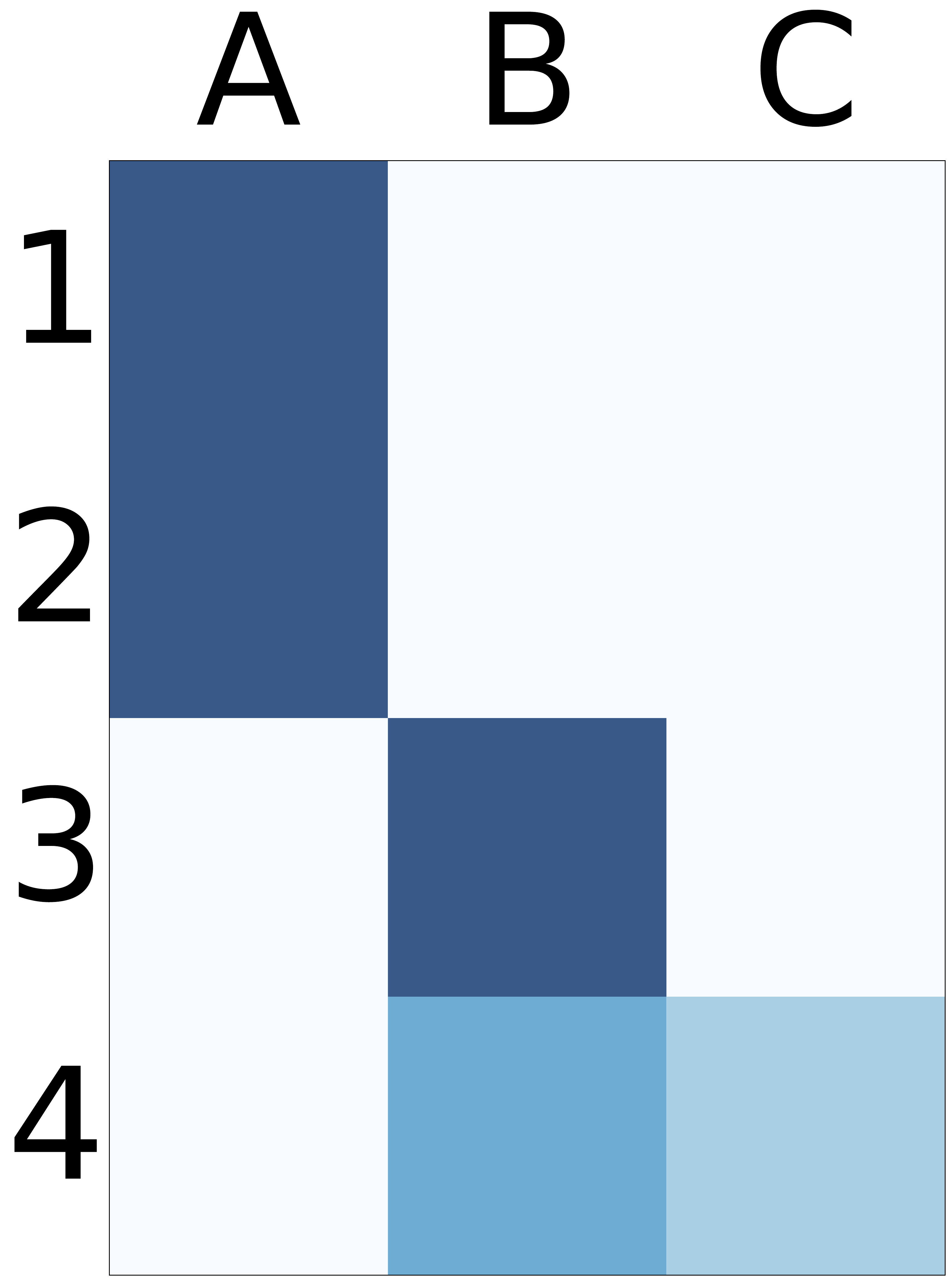}}$\Bigg)^{\text{T}}\quad\mathbf{\centerdot}\quad$
\raisebox{-.5\height}{\includegraphics[width=0.2\textwidth]{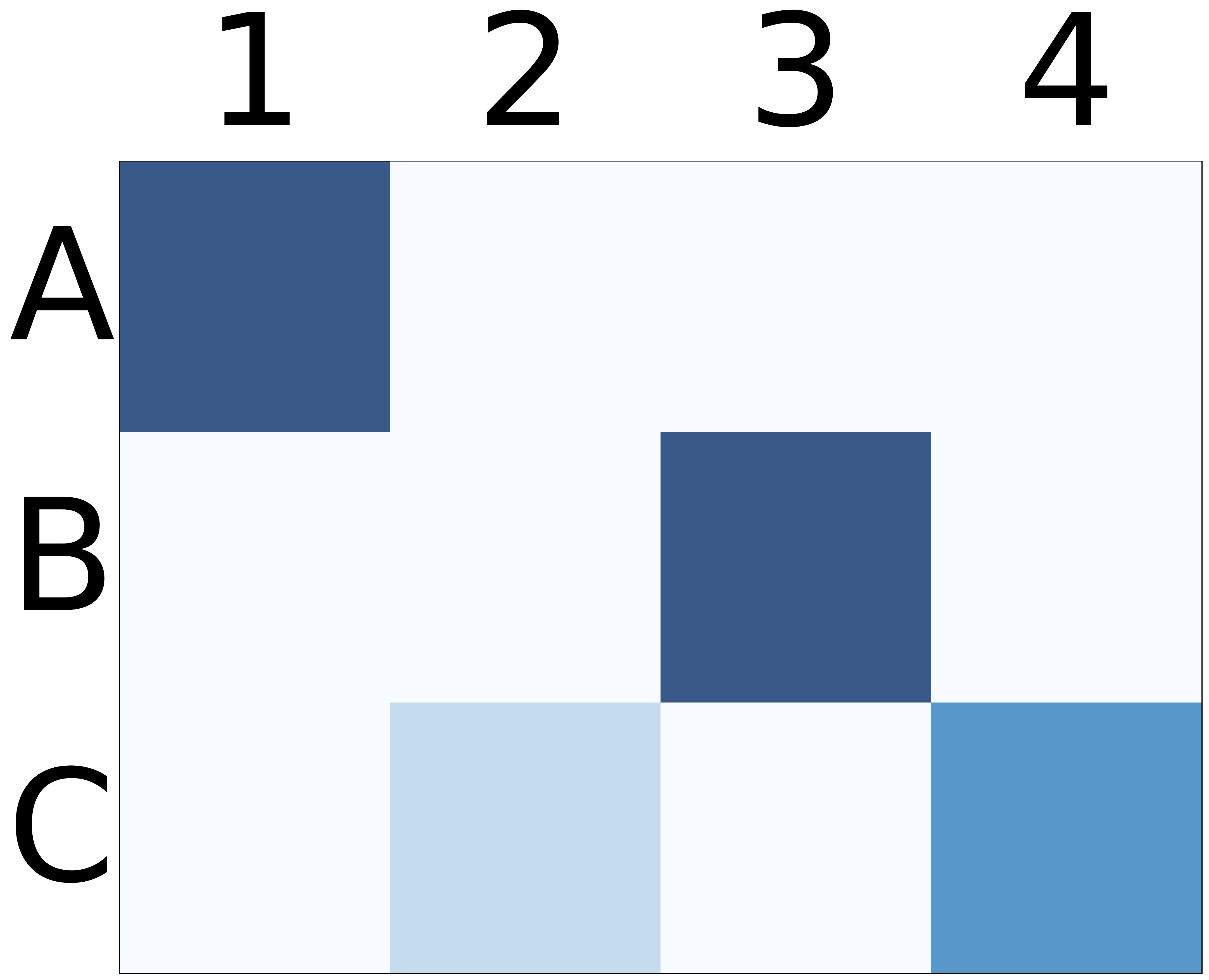}}$\quad=\quad$
\raisebox{-.5\height}{\includegraphics[width=0.2\textwidth]{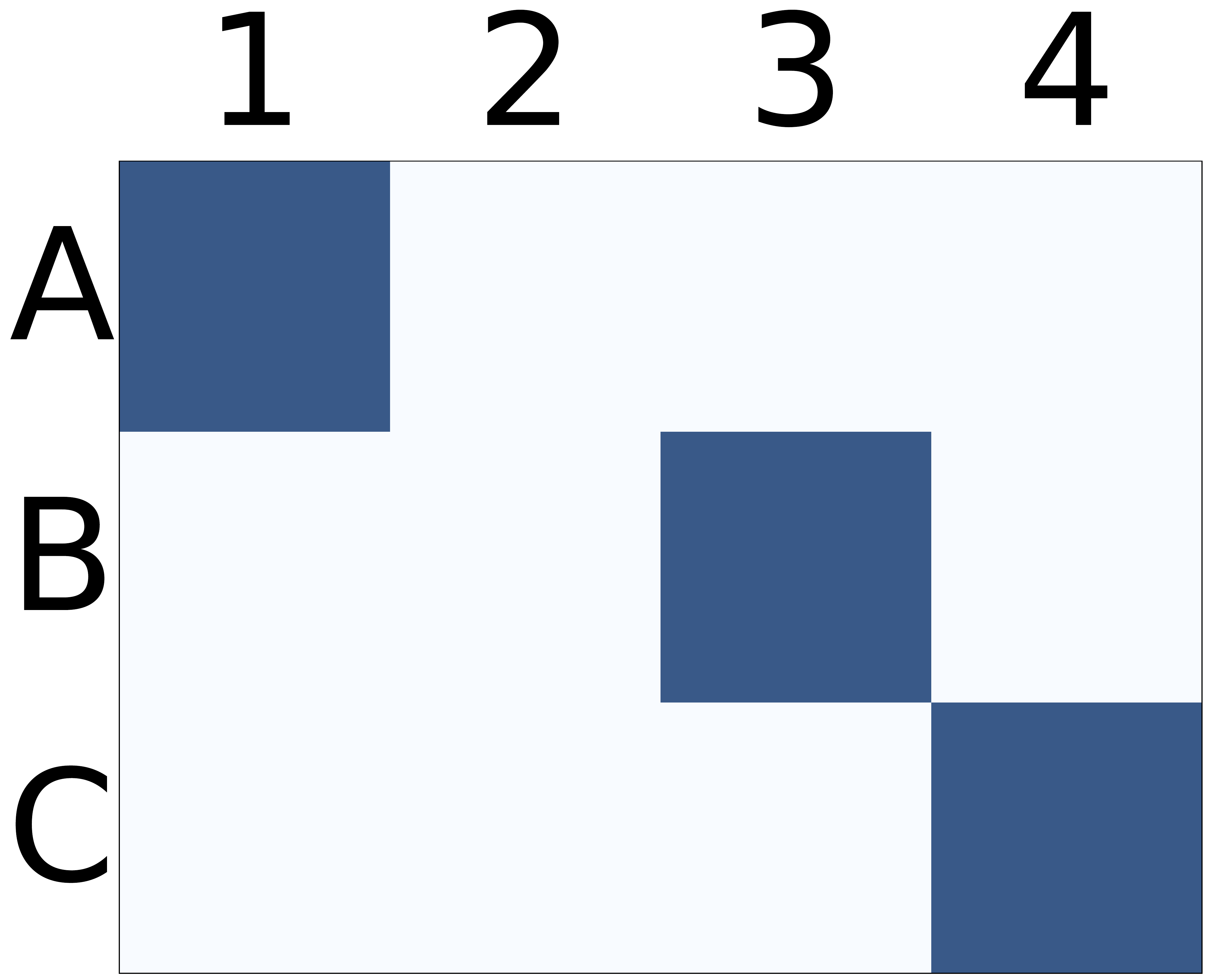}}
\end{center}
\caption{An example of dual-attention eliminating
uncertainty in the alignment.
In the left attention matrix, word ``4'' can be aligned to
either ``B'' or ``C''. In the middle attention matrix,
word ``C'' can be aligned to either ``2'' or ``4''.
The element-wise product eliminates these uncertainty 
and results in the right attention matrix.
\label{fig:dualatt}
}
\end{subfigure}
\vspace{-0.4cm}
\caption{Entailment composition (a) and dual-attention calculation (b).}
\vspace{-0.5cm}
\end{figure}

\iffalse
Note the resemblance between the above function
and the definition of Binary-Tree LSTM transition function
(Equation~\ref{eq:lstmdef}),
this suggests that we can directly use a Binary-Tree LSTM
layer to calculate the entailment 
in a way similar to \newcite{wang2015learning}.
That is, using $[\vech_i; \sum_{j\in P}\mtxa_{i,j} \vech_j]$
as the input, and $\vece_{i, 1}$, $\vece_{i, 2}$ as 
the hidden states passed from children.\footnote{
We also need to add two vectors representing the memories
from the children.
}
\fi
Figure~\ref{fig:entnode} illustrates the calculation
of the entailment composition.
%We have various choices for the composition function $f(\cdot)$
%in Equation~\ref{eq:relcomp}.
We will discuss $f_{\text{rel}}$ in Section~\ref{sec:treelstm}.

%%%%%%%%%%%%%%%%%%%%%%%%%%%%%%%%%%%%%%%%%%%%%%%%%%%%%%

\subsection{Dual-attention Over Tree Nodes}
\label{sec:dual}
\iffalse
\begin{figure}
\begin{center}
$\Bigg($\raisebox{-.4\height}{\includegraphics[width=0.08\textwidth]{figures/dualalign/left.pdf}}$\Bigg)^{\text{T}}\quad\mathbf{\centerdot}\quad$
\raisebox{-.5\height}{\includegraphics[width=0.1\textwidth]{figures/dualalign/right.pdf}}$\quad=\quad$
\raisebox{-.5\height}{\includegraphics[width=0.1\textwidth]{figures/dualalign/dual.pdf}}
\end{center}
\caption{An example of dual-attention eliminating
uncertainty in the alignment.
In the left attention matrix, word ``4'' can be aligned to
either ``B'' or ``C''. In the middle attention matrix,
word ``C'' can be aligned to either ``2'' or ``4''.
The element-wise product eliminates these uncertainty 
and results in the right attention matrix.
\label{fig:dualatt}
}
\vspace{-0.4cm}
\end{figure}
\fi

We can further improve our alignment approximation 
in Section~\ref{sec:att},
which does not consider any structural information of 
current tree, nor any alignment information from the premise
tree.

We can take a closer look at our conceptual example in 
Figure~\ref{fig:egtree}.
Note that the alignments have, to some extent, 
a symmetric property: 
if a premise node $j$ is most relevant to a hypothesis node
$i$, then the hypothesis node $i$ should also be most relevant
to premise node $j$.
For example, in Figure~\ref{fig:egtree},
the premise phrase ``hugging one another'' contradicts
the hypothesis word ``sleeping''.
In the perspective of the premise tree,
the hypothesis word ``sleeping''
contradicts by the known claim ``hugging one another''.
This suggests us to calculate the 
alignments from both side, and
eliminate the unlikely alignment
if it only exists in one side.
This technique is similar to the 
widely used forward and reversed alignment
technique in the machine translation area.

In detail, we calculate the expected alignments
$\tilde{\mtxa}$ from hypothesis to premise,
and also the expected alignments $\tilde{\mtxa}^R$
from premise to hypothesis,
and use their element-wise product
\[\tilde{\mtxa}^* = \tilde{\mtxa}\cdot\tilde{\mtxa}^R\] 
as the attention 
to feed into the Entailment Composition module.\footnote{
We need to normalize $\tilde{\mtxa}^*$ at each row to make
each row a probability distribution.
} This element-wise product is a mimic 
of the intersection %operation 
of two alignments in machine translation.
Figure~\ref{fig:dualatt} shows an example. %of dual-attention.

In addition to our dual-attention, 
\newcite{cohn2016incorporating} also explore to use the
structural information to improve the alignment.
However, their approach requires introducing 
some extra terms in the objective function, 
and is not straightforward
to integrate into our model. 
We leave adding more structural constraints
to further improve the attention 
as an open problem to explore in the future.

\iffalse
\section{Structured Tree Entailment}
\label{sec:entailment}
\input{entailment}
\fi

\section{Review: Recursive Tree Meaning Representations}
\label{sec:treelstm}
%!TEX root = main.tex

%We assume the node-level meaning representations over the
%premise tree and the hypothesis tree are provided.
%In this paper we use the Tree-LSTM model \cite{tai2015improved}
%to generate this representation. 
%However, our model is independent to this meaning representation
%and any high-quality meaning representation is suffice 
%for us to infer the alignments and entailments.

Here we describe the final building block of our neural model.

In Section~\ref{sec:att}, we did not mention the calculation of the
meaning representation $\vech_i$ for node $i$ in Equation~\ref{eq:att},
which represents the semantic meaning of the subtree rooted at node $i$.
In general, $\vech_i$ should be calculated recursively from the meaning
representations $\vech_{i,1}$, $\vech_{i,2}$ of its two children 
if node $i$ is an internal node, otherwise $\vech_i$ should be calculated
based on the word $\vecx\in\mathbb{R}^d$ in the leaf. 
\begin{equation}
\vech_i = f_{\text{MR}}(\vecx_i, \vech_{i,1}, \vech_{i, 2}).
\label{eq:mr}
\end{equation}

Similar is Equation~\ref{eq:relcomp}, where the relation $\vece_i$
is recursively calculated from the relation of its two children, 
as well as the meaning $\vech_i$
comparing with the meaning of the premise tree:
\begin{equation}
\vece_i
= f_{\text{rel}}([\vech_i; \sum_{j\in P}\tilde{\veca}_{i,j} \vech_j], \vece_{i, 1}, \vece_{i, 2}).
\label{eq:rel}
\end{equation}

Note the resemblance between these two equations, which indicates that
we can handle them similarly with the same form of composition function
$f(\cdot)$.

We have various choices for composition function $f$. For example,
we can use simple RNN functions as in \newcite{socher2013recursive}.
Alternatively, we can use a convolutional layer to extract features
from $\vecx_i, \vech_{i, 1}, \vech_{i, 2}$ and use pooling as aggregation 
to form $\vech_i$. 
In this paper we choose Tree-LSTM model \cite{tai2015improved}.
Our model is independent to this composition function
and any high-quality composition function is sufficient 
for us to infer the meaning representations and entailments.

Here we use Equation~\ref{eq:mr} as an example. Equation~\ref{eq:rel}
can be handled similarly.
Similar to the classical LSTM model \cite{hochreiter1997long}, 
in the binary Tree-LSTM model of \newcite{tai2015improved}, 
each tree node has a state represented by a pair of vectors:
the output vector $\vech\in\mathbb{R}^{1\times k}$, 
and the memory cell $\vecc\in\mathbb{R}^{1\times k}$,
where $k$ is the length of the Tree-LSTM
output representation. We use $\vech$ as the meaning representation
of the tree node in the attention model.
The LSTM transition calculates
 the state $(\vech_i, \vecc_i)$ of node $i$ 
with leaf word $\vecx_i\in \mathbb{R}^d$,
and two children with states $(\vech_{i, 1}, \vecc_{i,1})$
and $(\vech_{i,2}, \vecc_{i,2})$ respectively.

We can abuse the mathematics a little bit, 
and write the transition at an LSTM unit as a function:
\[
[\vech_i; \vecc_i] = \mathrm{LSTM}(\vecx_i, [\vech_{i,1}; \vecc_{i,1}], [\vech_{i,2}; \vecc_{i,2}])
\label{eq:lstmdef}
\]
In practice, we use the above $\mathrm{LSTM}(\cdot, \cdot, \cdot)$ function as $f_{\text{MR}}(\cdot, \cdot, \cdot)$, and $f_{\text{rel}}(\cdot, \cdot, \cdot)$. But we only expose the output $\vech_i$ 
to the above layers, and keep the memory $\vecc_i$ visible only
to the $\mathrm{LSTM}(\cdot, \cdot, \cdot)$ function.

Following \newcite{zaremba2014recurrent}, 
function $\mathrm{LSTM}(\cdot, \cdot, \cdot)$ is summarized by 
Equations~\ref{eq:treelstm-input}-\ref{eq:treelstm-hidden}:
\begin{equation}
\left(
\begin{array}{c}
\veci_i\\
\vecf_{i,1}\\
\vecf_{i,2}\\
\veco_{i}\\
\vecu_{i}
\end{array}
\right) = \left(
\begin{array}{c}
\sigma\\
\sigma\\
\sigma\\
\sigma\\
\tanh
\end{array}
\right) T_{d+2k, k}
\left(\begin{array}{c}
\vecx_i\\
\vech_{i,1}\\
\vech_{i,2}
\end{array}
\right)
\label{eq:treelstm-input}
\end{equation}
\vspace{-0.7cm}
\begin{align}
\vecc_i & = \veci_i \odot \vecu_i + \vecf_{i, 1} \odot \vecc_{i, 1}+ \vecf_{i, 2} \odot \vecc_{i, 2},\label{eq:treelstm-mem}\\
\vech_i & = \veco_i \odot \tanh(\vecc_i), \label{eq:treelstm-hidden}
\end{align}
where $\veci_i$, $\vecf_{i,1}$, $\vecf_{i,2}$, $\veco_i$ represent the
input gate, two forget gates for two children nodes, 
and the output gate respectively.
$T_{d+2k,k}$ is an affine transformation from $\mathbb{R}^{d+2k}$
to $\mathbb{R}^k$.
%In our practice, for leaf nodes,
%both $(\vech_{j, \ell}, \vecc_{j,\ell}), \ell=1,2$ are pairs
%of zero vectors.

%The above transition equations can be summarized by 
%Figure~\ref{fig:treelstm} \cite{tai2015improved}.

\newcolumntype{H}{>{\setbox0=\hbox\bgroup}c<{\egroup}@{}}
\begin{table*}
\begin{center}
\small
\begin{tabular}{c|cccHc}
Method & $k$ & $\card{\theta}_M$ & Train & Dev. & Test\\
\hline\hline
LSTM sent. embedding \cite{bowman2015large} 
& 100 & 221k & 84.8 & - & 77.6\\ \hline
Sparse Features + Classifier \cite{bowman2015large} 
& - & - & 99.7 & - & 78.2\\ \hline
%LSTM w/ sentences concatenation \cite{rocktaschel2015reasoning} & 
%116 & 252k &83.5 & 82.1 & 80.9\\ \hline
LSTM + word-by-word attention \cite{rocktaschel2015reasoning}
& 100 & 252k & 85.3 & 83.7 & 83.5\\ 
\hline
%mLSTM \cite{wang2015learning} & 150 & 544k &91.0 & 86.2 & 85.7\\ 
%\hline
mLSTM \cite{wang2015learning} & 300 & 1.9m &92.0 & 86.9 & 86.1\\ 
\hline
LSTM-network \cite{cheng2016long} & 450 & 3.4m & 88.5 & - & 86.3\\
\hline \hline

LSTM sent. embedding (our implement. of \newcite{bowman2015large})&100& 241k &79.0 & 78.9 & 78.4 \\ \hline
Binary Tree-LSTM (our implementation of \newcite{tai2015improved})& 100& 211k &82.4 & 80.2 & 79.9\\ \hline
Binary Tree-LSTM + simple RNN w/ attention & 150 & 220k &82.4 & 82.1 & 81.8\\
\hline
\hline
Binary Tree-LSTM + Structured Attention \& Composition & 150 & 0.9m & 87.0 & {\bf 87.3} & 86.4\\
\hline 
+ dual-attention & 150 & 0.9m & 87.7 & 87.1 & {\bf 87.2}
\end{tabular}
\end{center}
\vspace{-0.2cm}
\caption{Comparison between our structured model with other existing methods.
Column $k$ specifies the length of the meaning representations.
$\card{\theta}_M$ is the number of parameters
without the word embeddings.
\label{tab:results}}
%\vspace{-0.5cm}
\end{table*}

\iffalse
\begin{figure}
\centering
\includegraphics[width=0.4\textwidth]{figures/treelstm.pdf}
\caption{Transitions in one Tree-LSTM unit \protect\cite{tai2015improved}.
The parent node is denoted with subscript $j$,
and the children node with $j,1$ and $j,2$.
At each node, $\vecc$ represents the memory cell, 
$\vech$ represents the output,
and $\vecx$ represents the optional meaning representation of the word.
Some edges are labeled with the corresponding gates 
controlling the propagation of the information.
\label{fig:treelstm}}
\vspace{-0.4cm}
\end{figure}
\fi

\section{Empirical Evaluations}
\label{sec:exp}
%!TEX root = main.tex

We evaluate the performances of our structured attention model
and structured entailment model on the Stanford
Natural Language Inference (SNLI) dataset \cite{bowman2015large}.
The SNLI dataset
contains $\sim570$k sentence pairs.
%Each pair is labeled with one of the relations of
%entailment, neutral, contradiction and $-$,
%where $-$ means there is no consensus among the annotators.
%All those sentence pairs with $-$ labels are discarded.
We use the binarized trees in SNLI dataset in our experiments.

\begin{figure*}[ht]
\begin{center}
\begin{tabular}{cc}
\raisebox{0pt}{\includegraphics[width=0.35\textwidth]{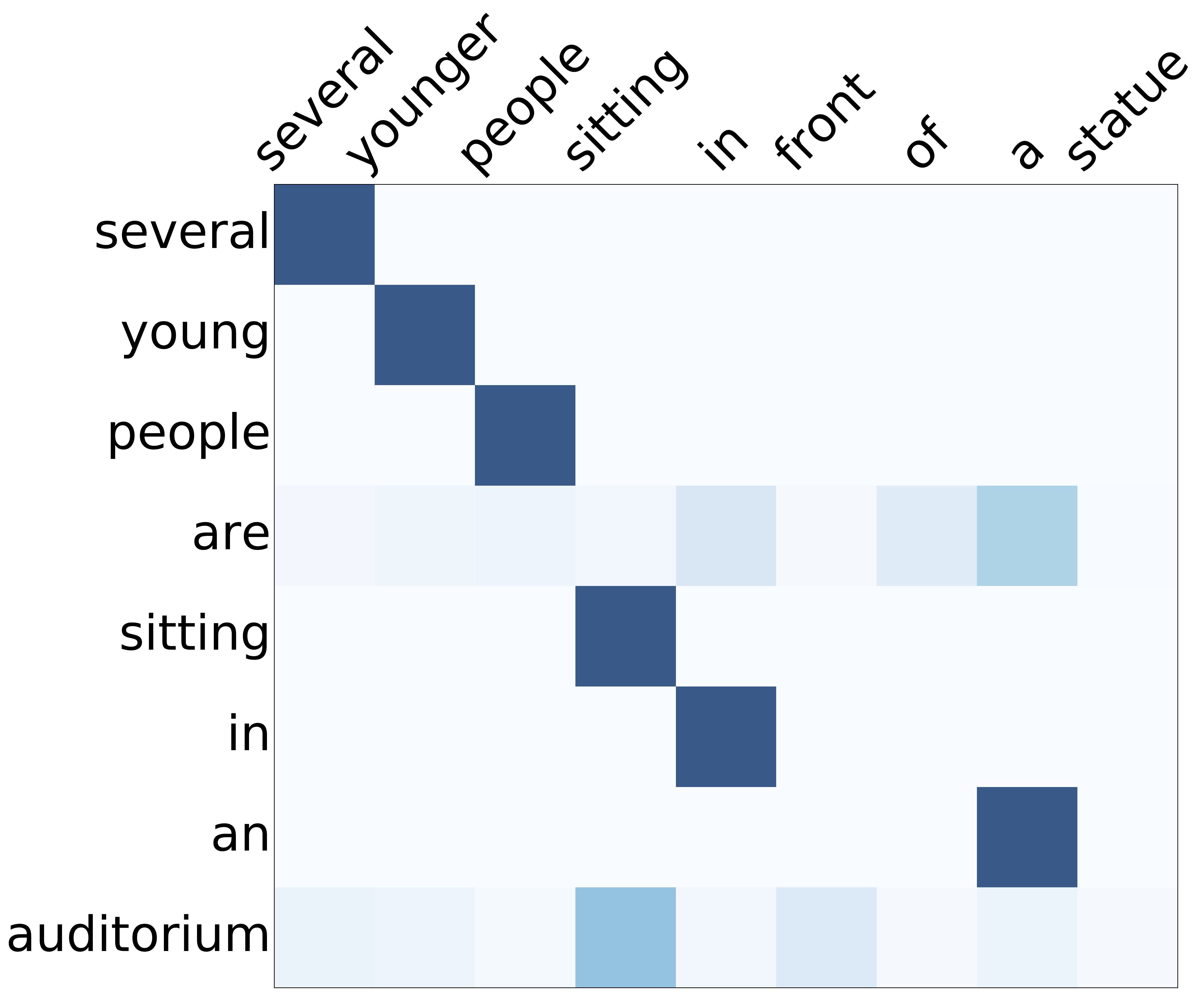}} &
\includegraphics[width=0.35\textwidth]{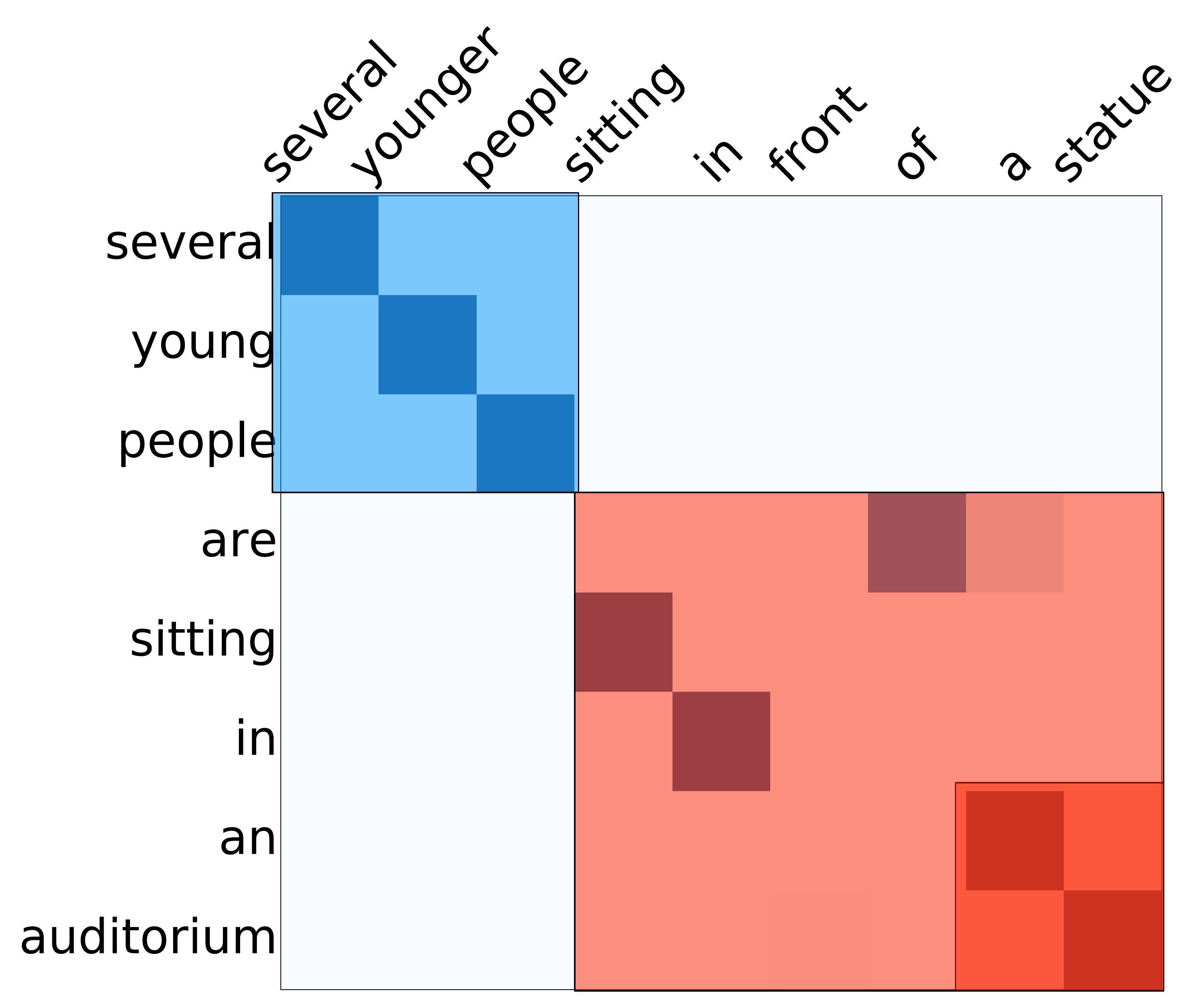} \\
(a) attention & (b) dual-attention\\
\raisebox{0pt}{\includegraphics[width=0.35\textwidth]{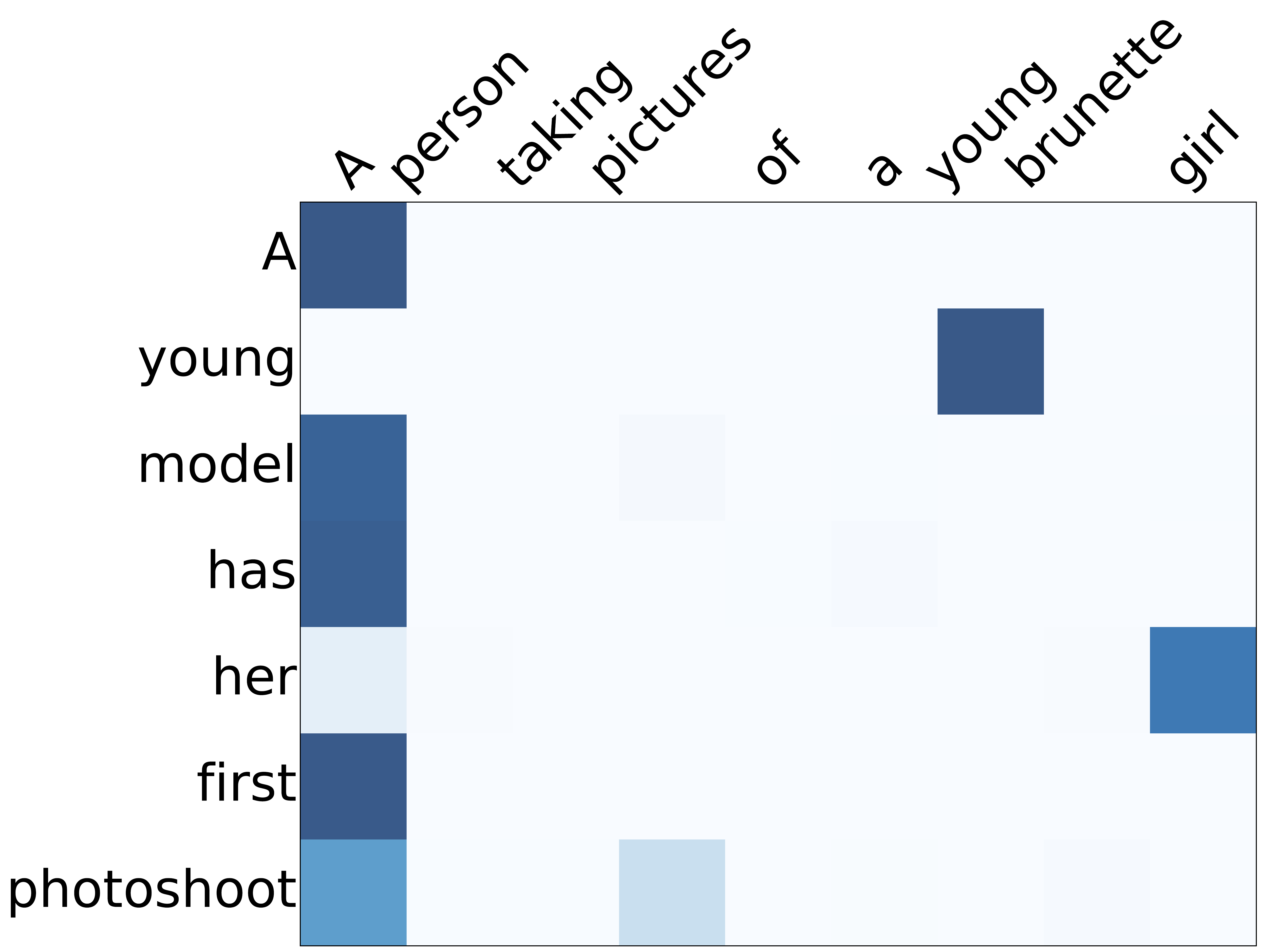}} &
\includegraphics[width=0.35\textwidth]{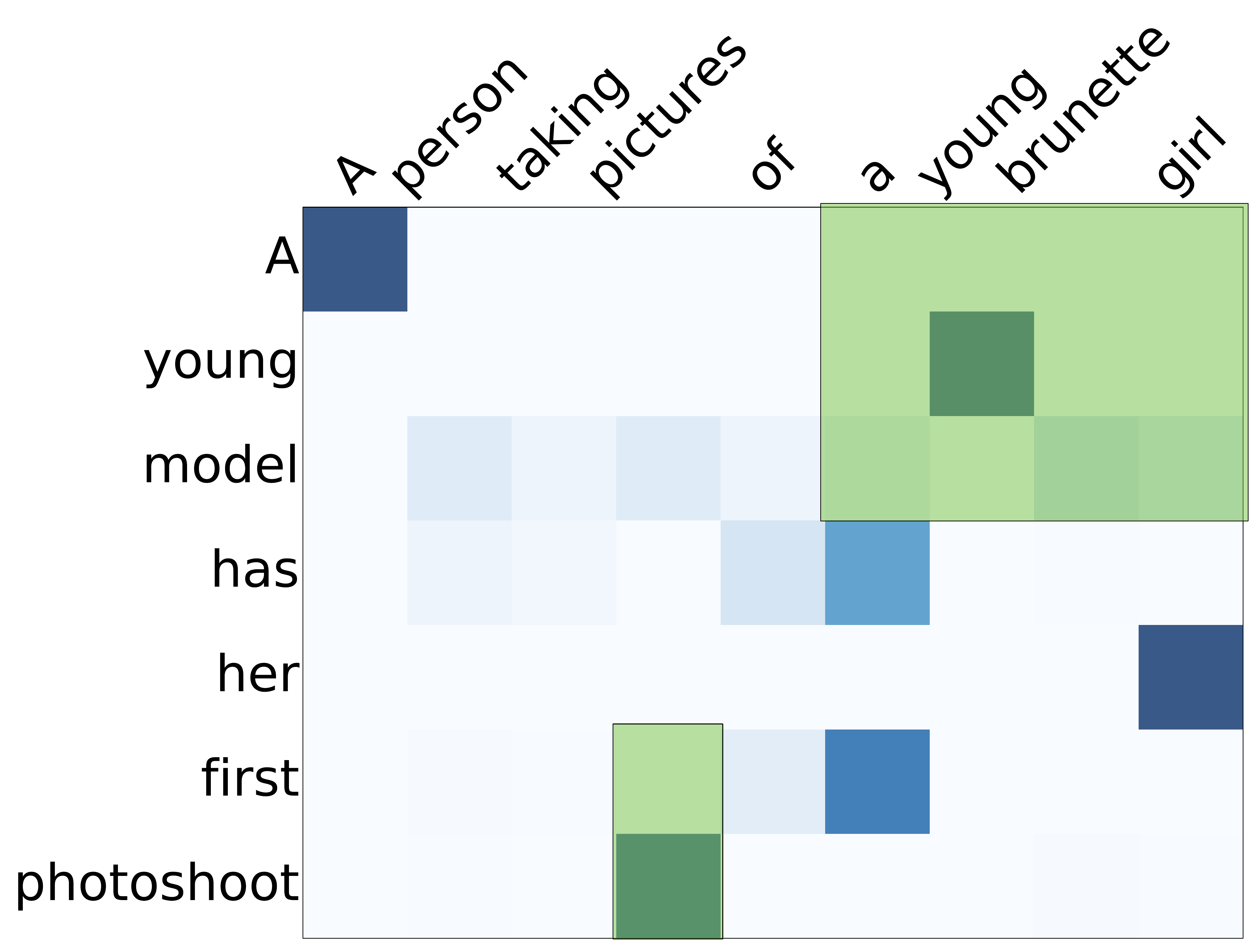} \\
(c) attention & (d) dual-attention\\
\end{tabular}
\end{center}
%\vspace{-0.2cm}
\caption{Attention matrices for exemplary sentence pairs.
Note that, for brevity we {\it only show} the attentions 
between each word pair, and skip
the attentions of tree nodes.
Some important tree node alignments calculated by our model are highlighted 
using the colored boxes,
where the colors of the boxes represent the entailment relations (see
Figure~\ref{fig:ent-example}).
(a) (b) Premise: several younger people sitting in front of a statue. Hypothesis: several young people sitting in an auditorium.
Dual-attention fixes the misaligned word ``auditorium''.
(c) (d) Premise: A person taking pictures of a young brunette girl.
Hypothesis: A young model has her first photoshoot. 
Dual-attention fixes the uncertain alignments for ``photoshoot''
and ``model''.
\label{fig:align-example}}
%\vspace{-0.5cm}
\end{figure*}

\begin{figure*}[ht]
\begin{center}
\begin{tabular}{c}
%\vspace{-0.2cm}
\raisebox{1.5in}{(a)}
\includegraphics[width=0.97\textwidth]{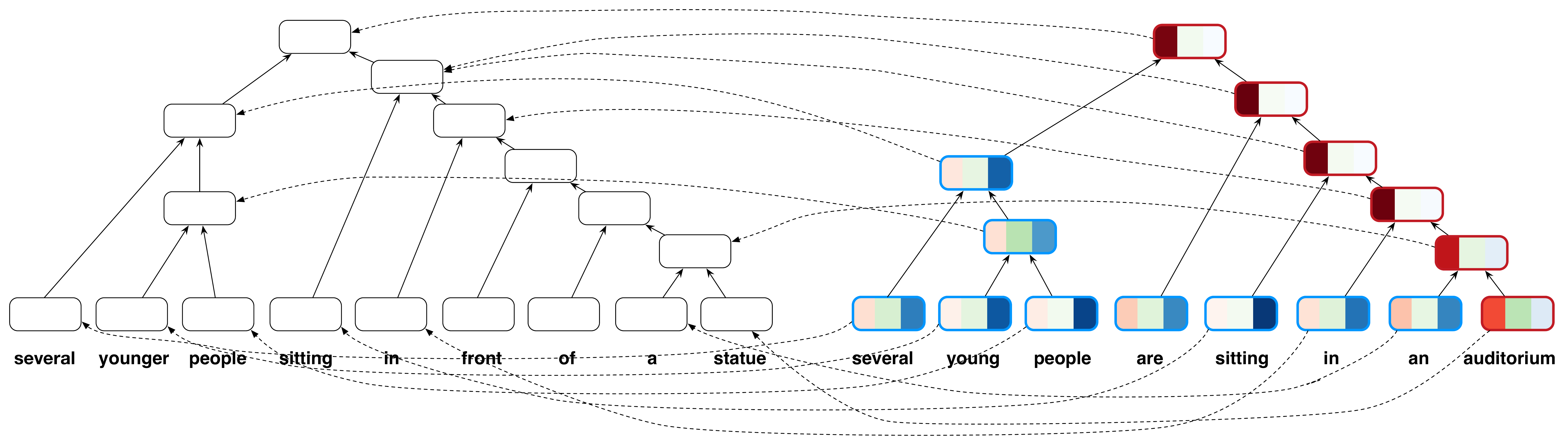}\\
%\vspace{-0.1cm}
%(a) \\
\raisebox{1.5in}{(b)}
\includegraphics[width=0.97\textwidth]{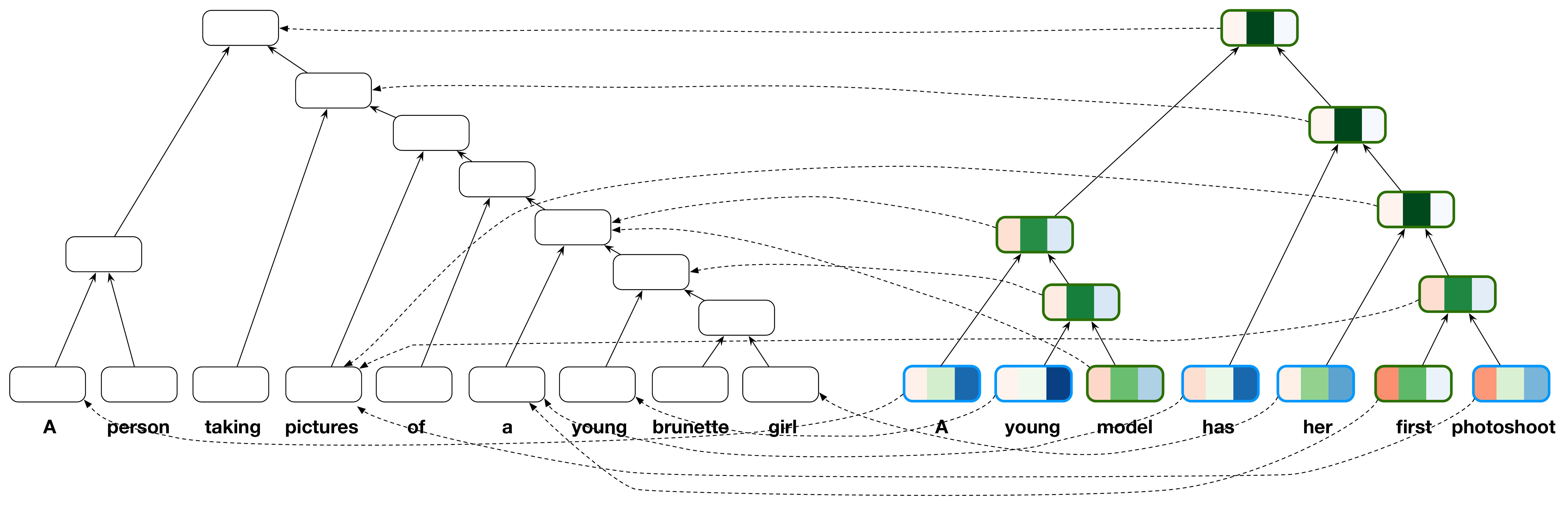}% \\
%\vspace{-0.2cm}
%(b)
\end{tabular}
\end{center}
%\vspace{-0.5cm}
\caption{Examples illustrating entailment relation composition.
(a) for Figure~\ref{fig:align-example} (b); (b) for Figure~\ref{fig:align-example} (d). 
For each hypothesis tree node, the dashed line shows
to its most confident alignment.
The three color stripes in each node indicate
the confidences of the corresponding entailment relation
estimation: red for {\color{red} contradiction}, 
green for {\color{green} neutral}, and blue for {\color{blue} entailment}.
The colors of the node borders show the dominant 
estimation.
Note: there is no strong alignment for hypothesis word ``are'' in (a).
\label{fig:ent-example}}
%\vspace{-0.4cm}
\end{figure*}

%\vspace{-0.3cm}
\subsection{Experiment Settings}

{\bf Network Architecture}

The general structure of our model is illustrated 
in Figure~\ref{fig:arch}.
We omitted a dropout layer between the word embedding 
layers and the tree LSTM layers in Figure~\ref{fig:arch}.
%at the bottom there is a embedding layer that maps 
%each natural language sentence to a sequence
%of word embedding vectors. 
%The output of this layer is then passed to a dropout layer.
%After the dropout
%the result is fed into a Binary-Tree LSTM to generate
%meaning representations for each tree node.
%The Alignment Module (Section~\ref{sec:att})
%calculates the expected alignment (attention)
%based on the node meaning representation.
%The Entailment Composition Module (Section~\ref{sec:ent})
%uses the expected alignment and the tree node meaning representations
%to compute the entailment relation representation
%along the hypothesis tree from bottom up,
%and the final result is passed to a softmax layer
%to calculate the probabilities for each of the 3 relations.
We use cross-entropy as the training objective.\footnote{
Our code is released at \url{https://github.com/kaayy/structured-attention}.}

%We implement our system using {\tt torch}.

\noindent {\bf Parameter Initialization \& Hyper-parameters}

We use GloVe \cite{pennington2014glove} to initialize the 
word embedding layer. In the training we do not change the embeddings,
except for the OOV words in the training set.
For the parameters of the rest layers,
we use a uniform distribution between $-0.05$ and $0.05$
as initialization.

Our model is trained in an end-to-end manner with adam \cite{kingma2014adam} as the optimizer.
We set the learning rate to 0.001, $\beta_1$ to 0.9,
and $\beta_2$ to 0.999.
We use minibatch of size 32 in the training.
The dropout rate is 0.2.
The length for the Tree-LSTM meaning representation
$k=150$. The length of the entailment relation
vector $r=150$.

%\vspace{-0.3cm}
\subsection{Quantitative Evaluation}

We present a comparison of structured model with existing methods
of LSTM-based sentence embedding \cite{bowman2015large},
LSTM with attention \cite{rocktaschel2015reasoning},
Binary Tree-LSTM sentence embedding (our implementation
of \newcite{tai2015improved}),
mLSTM \cite{wang2015learning},
and LSTM-network \cite{cheng2016long} 
in Table~\ref{tab:results}.

%We start with introducing Binary-Tree LSTM for sentence 
%meaning representation to \cite{bowman2015large}. 
%This meaning representation
%is then feed directly into an NN similar to \cite{bowman2015large}.
%With this simple replacement of the sentence meaning representation
%brings $\sim$1.5 improvement to 79.9 in the test set,
%which reveals the potential of learning sentence representations
%with sentence structures.

We first try Binary-Tree LSTM with a composition function
$f_{\text{rel}}$ of a 
recurrent network with attention as 
in \newcite{rocktaschel2015reasoning},
which achieves an accuracy of 81.8.
%based on the Binary-TreeLSTM sentence embedding.
%By checking the training statistics 
%(e.g., the norm of the gradient passed between the
%network layers), we find that
%this network has the vanishing gradient issue,
%which might explain why it does not perform as well
%as the same RNN in the sequence model.
We find the training of this RNN is difficult due
to the vanishing gradient problem.

Using Binary-Tree LSTM for entailment relation composition
instead of the simple RNN brings $\sim$4.6 improvement. 
%We observe that the gradients passed between the network
%layers are usually 2 orders of magnitude larger 
%than the simple RNN model, which might explain
%the superior performance.
We observe that the vanishing gradient problem is greatly
alleviated.
Dual-attention further improves the tree
node alignment, achieving another 0.8 improvement.

%Comparing with other LSTM-based methods, %in the same task,
Our structured entailment composition model outperforms
the similar mLSTM model, which essentially
also uses an LSTM layer to propagate 
the ``matching'' information, but sequentially. %not
%in the bottom-up order.
With the help of dual-attention, our model outperforms
mLSTM with a 1.1 point margin.

%LSTM-network model \cite{cheng2016long}
%applies attention within the 
%sentence, 
%which is orthogonal to our work. 
%Ideally we can also integrate the intra-sentence
%attention to our model. We leave this as further work.

%%%%%%%%%%%%%%%%%%%%%%%%%%%%%%%%%%%%%%%%%%%
%\vspace{-0.3cm}
\subsection{Qualitative Evaluation}
\label{sec:qual-eval}

Due to space constraints, 
here we highlight two examples in Figure~\ref{fig:align-example} for both standard attention and dual-attention,
and Figure~\ref{fig:ent-example} for entailment composition.
To pick the most representative examples from the dataset needs
careful consideration.
Ideally random selection is most convincing. However,
due to the fact that most correctly classified examples 
in the datasets
are trivial sentence pairs with only word insertion, deletion, or replacement,
and many incorrectly classified examples in the datasets 
involves common knowledge, (e.g., ``waiting in front of a red light''
entails ``waiting for green light'',
or ``splashing through the ocean'' contradicts ``is in Kansas'',)
it is time-consuming to find meaningful insights
from randomly selected examples.
Here we manually choose two examples from the test set 
of the SNLI corpus,
with consideration of both generality 
and non-triviality. They both involve complex syntactic structures and
compositions of several relations. 
In addition, some examples that need more subtle linguistic insights 
are discussed in Section~\ref{sec:disc}.

Our first example is shown in Figure~\ref{fig:align-example} (a) and (b),
with premise ``several younger people sitting in front
of a statue'',
and hypothesis  
``several young people sitting in an auditorium''.
Figure~\ref{fig:align-example} (a) and (b) only show the
word-level attention for brevity.
In this example, note the hypothesis word ``auditorium'',
which has no explicit correspondence in the
premise sentence, but indeed has an implicit correspondence
``statue'' that indicates the conflict relation.
The standard attention model aligns ``auditorium'' to ``sitting''
since they more frequently co-occur,
leading to an incorrect relation of
``entailment'' (not shown in Figure~\ref{fig:ent-example}).
The dual-attention model correctly finds the alignment between
``auditorium'' and ``statue'' since ``sitting'' is more likely
to be aligned to the same word in the premise.
The colored boxes in Figure~\ref{fig:align-example} (b)
show some important tree node alignment calculated by our model. 
The colors
represent the entailment relation based on the alignment,
%which is composed recursively from bottom up,
as shown in Figure~\ref{fig:ent-example} (a).

In Figure~\ref{fig:ent-example} (a), each tree node is filled
with three color stripes, whose darknesses show the confidences
of the corresponding entailment relations.
For this example, the contradiction relation from ``statue''
and ``auditorium'' flips every tree node from bottom up
and finally make the final result contradiction,
%For this specific example, most parts of the sentence are considered
%entailment relations, but the final conclusion is flipped
%to contradiction, since ``statue'' contradicts ``auditorium''.
%This contradiction relation propagates from the rightmost
%leaf towards the root via the entailment composition,
similar to our concept example in Figure~\ref{fig:egtree}.

Another example with premise 
``a person taking pictures of a young brunette girl'',
and hypothesis 
``a young model has her first photoshoot''.
The word-level attentions are shown in Figure~\ref{fig:align-example}
(c) (d).
The standard attention is uncertain about two words:
1) word ``model'' has several meanings, making it hard to find 
the right alignment, but
in the perspective of from premise to hypothesis,
it is easier since a girl is more likely to be a model.
2) Similar is for hypothesis word ``photoshoot'',
which can either be aligned to ``a'' or ``pictures''
%in the hypothesis to premise perspective, 
but since ``a'' is aligned to other words,
dual-attention aligns ``photoshoot'' to ``pictures''.

In Figure~\ref{fig:ent-example} (b), we can see that
there are two parts in the hypothesis indicates that
the relation should be neutral: 1) ``a young brunette girl''
is not necessarily a ``a young model''; and 2)
the ``pictures'' taken are not necessarily ``her first photoshoot''.

\subsection{Discussion}
\label{sec:disc}

Although many attention-based
models, including our model, 
achieve superior 
results in the Stanford Natural Language Inference dataset,
we still need to circumvent some problems
to apply these neural models to more 
general textual entailment
problems.

Despite those sentence pairs that require
more common knowledge to find the entailment relations
as we mentioned in Section~\ref{sec:qual-eval},
we are more interested in sentences that are difficult
because they involve non-trivial linguistic properties.

Consider the following two pairs of sentences
that are difficult for current attention and composition based models:
\begin{enumerate}
\item \begin{itemize}
\item Premise: The boy loves the girl.
\item Hypothesis: The girl loves the boy.
\end{itemize}
Here the only difference between the two sentences
is the order/structure of the words. To handle this problem
the attention-based models should take the reordering
into consideration when composing entailment relations.
\item 
\begin{itemize}
\item Premise: A stuffed animal on the couch.
\item Hypothesis: An animal on the couch.
\end{itemize}
In this example, almost every hypothesis word occurs in the premise sentence,
but it is difficult to infer that ``a stuffed animal'' 
is not ``an animal''. While in most cases
the monotonicity of entailment suggests that
a word deletion in the premise sentence either leads
to entailment, e.g., ``a cute animal'' entails ``an animal'',
or a reverse entailment, e.g., ``some animal'' reverse entails
``animal'' (See \newcite{maccartney2009extended} for more details),
but for words like ``stuffed'' it is quite different:
their monotonicity directions depend on the nouns being modified,
e.g.,
``a stuffed animal'' does not entails ``an animal'', but ``a stuffed toy'' entails ``a toy''.
This observation suggests that
we might need to consider phrases like ``stuffed animal'' as a whole 
instead of treating the two words separately and then 
composing  the entailment relations.
\end{enumerate}

In addition, training of the neural models rely on large training corpora,
which makes it difficult to directly apply neural models on
traditional RTE datasets, e.g., the Pascal RTE dataset \cite{dagan2006pascal} and the FraCaS dataset \cite{cooper1996using}, which are usually small
and contain many named entities that are hard for neural models to identify.

\iffalse
\section{Discussion}
\label{sec:disc}
\input{disc}
\fi

\section{Conclusion}
\label{sec:conclusion}
We have presented an approach to model the composition
of the entailment relation following the tree structure for the sentence entailment task. We adapted the attention model
for tree structures. Experiments show that
our model bring significant improvements in accuracy,
and is easy to interpret.

\section*{Acknowledgments}
We thank the anonymous reviewers for helpful comments.
We are also grateful to James Cross, Dezhong Deng, and Lemao Liu for suggestions.
This project was supported in part
by NSF IIS-1656051, DARPA FA8750-13-2-0041
(DEFT), and a Google Faculty Research Award.

\bibliography{entailment}
\bibliographystyle{acl}

\end{document}